\documentclass[letterpaper, 10 pt, conference]{ieeeconf}  

\usepackage[ruled,vlined,noline,noend]{algorithm2e}
\usepackage{mathtools}
\usepackage{graphicx}
\usepackage{amsmath}
\usepackage{booktabs}
\usepackage{courier}
\usepackage{xcolor}
\usepackage{listings}
\usepackage{multirow}
\usepackage{float}
\usepackage[hidelinks]{hyperref}
\usepackage[utf8]{inputenc}
\usepackage{dsfont}
\usepackage{multicol}
\usepackage{subcaption}
\usepackage{balance}
\usepackage{comment}

\IEEEoverridecommandlockouts                              

\title{\LARGE \bf
Online Explanation Generation for Planning Tasks \\ in Human-Robot Teaming
}

\author{Mehrdad Zakershahrak, Ze Gong, Nikhillesh Sadassivam and Yu Zhang
\thanks{Mehrdad Zakershahrak, Ze Gong, Nikhillesh Sadassivam and Yu Zhang are with the School of Computing, Informatics and Decision Systems Engineering, Arizona State University, Tempe, AZ.
        {\tt\small \{mzakersh,zgong11,nsadassi,yzhan442\}@asu.edu} 
         {}
}%
}

\begin{document}

\newtheorem{definition}{Definition}

\maketitle
\thispagestyle{empty}
\pagestyle{empty}

\begin{abstract}
As AI becomes an integral part of our lives, the development of explainable AI, embodied in the decision-making process of an AI or robotic agent, becomes imperative. 
For a robotic teammate,
the ability to generate explanations to justify its behavior is one of the key requirements of explainable agency.
Prior work on explanation generation has been focused on  
supporting the rationale  
behind the robot's decision or behavior. 
These approaches, however, fail to consider the mental demand for understanding the received explanation.
In other words, the human teammate is expected to understand an explanation no matter how much information is presented.
In this work, 
we argue that explanations, especially those of a complex nature, should be made in an online fashion during the execution,
which helps spread out the information to be explained and thus reduce the mental workload of humans in highly cognitive demanding tasks. 
However, a challenge here is that the different parts of an explanation may be dependent on each other, which must be taken into account when generating online explanations. 
To this end, a general formulation of online explanation generation is presented with three variations satisfying different \textit{``online''} properties. 
The new explanation generation methods are based on a model reconciliation setting introduced in our prior work.
We evaluated our methods both with
human subjects in a simulated rover domain, using NASA Task Load Index (TLX), and synthetically with ten different problems across two standard IPC domains.
Results strongly suggest that our methods generate explanations that are perceived as less cognitively demanding and much preferred over the baselines and are computationally efficient.  
\end{abstract}

\section{INTRODUCTION}

As intelligent robots become more prevalent in our lives, the interaction of such AI agents with humans becomes more frequent and essential. 
One of the most important aspects of human-robot interaction is for the robotic agent to provide explanations to support the rationale behind its decision or behavior \cite{lombrozo2006structure}. 
An explanation provides justifications for the robot, which helps the human maintain trust of the robotic peer as well as a shared situation awareness \cite{endsley1988design,cooke2015team}. 
Prior work on explanation generation, however, often ignores the underlying requirements of the human recipient to understand an explanation \cite{gobelbecker2010coming,hanheide2017robot,sohrabi2011preferred}. 
A good explanation should be generated in a lucid fashion from 
the recipient's perspective~\cite{chakraborti2017plan,miller2018explanation}, so that it is understood.

To address this problem, 
a key consideration is that the human recipient ({\it explainee}) may interpret an explanation differently from the robot ({\it explainer}) due to a different understanding of the domain. 
In our prior work~\cite{chakraborti2017plan}, we refer to such differences as \textit{model differences}. 
The robotic agent, as a result, must ensure that the explanation makes sense in the human's model,
which generates the human's expectation of the robot,
so that the robot's behavior matches with the human's expectation.
An explanation can then be considered as a request to change the human's model to reduce the model differences so that the robot's behavior is consistent with the updated human model.
The decision-making process (including explanation generation discussed herein) in the presence of such model differences is more generally referred to as \textit{model reconciliation}~\cite{chakraborti2017plan, zhang2017plan}. 

One remaining challenge, however, is the consideration of the mental demand of the human to understand an explanation. 
In most prior work on explanation generation, the human is expected to understand an explanation regardless of how much information it contains. Little discussion has been given on the ways of presenting such information. In this work, we argue that explanations, especially complex ones, should be provided in an online fashion,
such that each explanation is broken into multiple parts, which are then communicated separately and intertwined with plan execution.
Communicating an explanation in such a manner is expected to result in less mental workload for cognitively demanding tasks since the information is spread out so that the interpretation process becomes incremental, which is known to benefit understanding~\cite{fischer1979styles}.
One of the main challenges here is that the different parts of an explanation could be dependent on each other, which must be taken into account when generating online explanations.
Our online explanation generation process spreads out the information while ensuring that the different parts do not introduce cognitive dissonance so that they are always perceived in a cohesive fashion. 

\subsection{Motivating Example} 
Let us illustrate the motivation of online explanation generation via a familiar situation involving a daily routine. Mark works at a company and has a voice assistant helping him get ready for work everyday. Usually Mark wakes up, drinks a freshly made coffee, enjoys a filling breakfast, dresses for work, and then drives to work. However, today, Mark has a meeting scheduled in the early morning so Mark needs to arrive at work earlier, a presentation at near lunch time, his car is broken, and there is no coffee beans for fresh coffee. The voice assistant knows that Mark must be reminded of these changes. However, explaining all the changes at the same time may result in unnecessary strain on Mark. In contrast, the robot first suggests Mark to prepare an instant coffee,, explaining that there is no coffee beans left. As Mark is enjoying his coffee, it reminds him to cook a light breakfast since there is an important meeting scheduled early today. As Mark is enjoying his breakfast, the voice assistant advises Mark to prepare a lunch box, since there is a presentation at near lunch time so that Mark may not have time to eat outside. After Mark is done with the lunch box, it asks Mark to call the taxi company since his car is broken. After breakfast, the assistant mentions that Mark needs to dress up today for work. When the taxi arrives, the voice assistant asks Mark to take the lunch box. Comparing the voice assistant’s strategy to share all of the information at the beginning, we can see that conveying the information in an online fashion is more cognitively friendly (i.e., involving less information at a time) and hence helps with reducing strain and cognitive load~\cite{kahneman2011thinking,cowan2008differences}.
These effects are highly desirable for tasks that are
cognitively demanding for humans. 

In this paper,
we develop a general formulation of online explanation generation by breaking an explanation up into multiple parts to be communicated at different times during plan execution.
We develop three variations of online explanation generation methods with each satisfying different ``online'' requirements. In the first method, the focus is for a robot to explain only plan prefixes.
This is in contrast to prior offline methods where the entire plan must be explained, which allows us to break an explanation up into multiple parts with each explaining only a part of the plan. 
We use a model search method to ensure that
the earlier parts communicated do not affect the latter parts of an explanation.
In the second method, we further relax the online requirement 
by requiring only the very next action to be explained (if needed).
The assumption here is that the actions already occurred do not affect the understanding of the robot's future actions,
which holds in situations where each action is viewed independently or the human has a short cognitive span (such as in highly demanding tasks).  
In the third method, we relax the assumption of the uniqueness of the human's interpretation of a plan and 
the robot is only required to explain with respect to any such interpretation.  
A compilation method is developed that converts this problem into one that requires solving two planning problems.  
Our methods are evaluated both synthetically and with human subjects in standard planning domains. 
Results strongly suggest that our methods not only generate explanations that are perceived as less cognitively demanding and much preferred over the baselines but also are computationally efficient. 

\section{Related Work}
\vskip-3pt
The advancement of AI and its numerous applications have provided astounding benefits in many areas such as transportation, medicine, finance, education, and entertainment. And yet AI agents have thus far been limited in their ability to 
operate as a teammate. 
To be considered a teammate, an AI agent must not only achieve a given task, but also provide a level of transparency about itself to other members of the team~\cite{cooke2015team}. 
One way to achieve this is to enable AI agents to be self-explanatory in their behaviors. 
Recently, the explainable AI paradigm \cite{gunning2017explainable} rises as one essential constituent of AI systems. Explainable AI maintains a shared situation awareness by facilitating the human's understanding of the AI agent, which also improves the human's trust.

The effectiveness of explainable agency \cite{langley2017explainable} depends on the agent's ability to model the human's interpretation of its behavior.
While there exists prior work that focuses on aligning the values \cite{brewka1996artificial} or goals \cite{andersen2016projecting, dragan2013legibility}, the interpretation also depends on the domain model~\cite{gong2020you}. 
This means that an explainable AI agent must not only model the domain, but also the human's interpretation of the domain~\cite{chakraborti2017ai}, which may be quite different.
This interpretation model of the human enables
the AI agent to infer about the human's expectation of itself. 
Using such a model, an agent can 
generate
explicable plans~\cite{zhang2017plan, zakershahrak2018interactive, fox2017explainable},
assistive actions \cite{reddy2018you}, etc., to facilitate fluent human-robot interactions.
In these methods, the AI agent substitutes a cost metric with 
a new metric that simultaneously considers the cost and a distance metric between the robot's behavior and the human's expectation of it.
Optimizing this new metric often leads to a trade-off between the plan cost and plan interpretability.

Another way to use the human's interpretation of the domain model requires explicit communication,
which has the benefit of maintaining cost optimality.
The model can be used to infer about which actions in an optimal robot plan are likely to introduce misinterpretations. 
In some cases, simply providing the future context for those actions 
is sufficient~\cite{gong2018robot} to make them interpretable.
Methods for analyzing the domain to identify the ``{\it causes}'' of the robot's plan (or its failures) have been studied before~\cite{gobelbecker2010coming,hanheide2017robot,sohrabi2011preferred}. 
These methods however assume no differences between the robot's and human's models.  
More recently, research work has been proposed to specifically address this issue by considering their differences and generating explanations to
reduce them~\cite{chakraborti2017plan, miller2018explanation}. 
However, all research above has been focused on generating the ``{\it right}'' explanation  while ignoring the cognitive requirement of the human for understanding the explanation.
In our prior work, we have studied how the ordering for presenting the information of an explanation may influence its interpretation~\cite{zhang2019progressive}.
In this work, we further argue that an explanation should be made in an online fashion for cognitively demanding tasks. 

\section{Explanation Generation as \\ Model Reconciliation}
\begin{figure}
\centering
    \includegraphics[scale = 0.4]{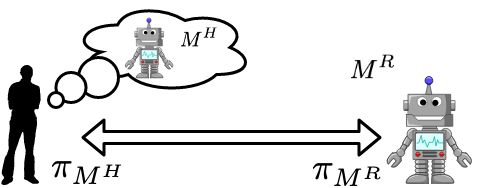}
    \caption{The model reconciliation setting first introduced in~\cite{chakraborti2017plan}. $M^R$ represents the robot's model and $M^H$ the human's interpretation model of the robot's behavior. Using $M^H$, the human obtains his expectation of the robot's behavior $\pi_{M^H}$. Whenever that is inconsistent with the robot's actual behavior $\pi_{M^R}$ (generated by $M^R$), the robot explains by generating  an explanation to reconcile the two models.}
    \label{fig1}
    \vspace{-1.5em}
\end{figure}

We consider the explanation generation problem in a model reconciliation setting first introduced
in our recent work~\cite{chakraborti2017plan}. 
The reason for this choice is that it represents a more general setting for explanation generation than those used in the previous work, which considers both the robot's ({\it explainer}) and human's ({\it explainee}) models as discussed in the related work.
An illustration of the model reconciliation setting is presented in 
Fig. \ref{fig1}. 
Next, we provide a brief review of the formulations used in our setting. 

Model reconciliation defines a planning setting. 
A planning problem is defined as a tuple $(F, A, I , G)$ using PDDL~\cite{fox2003pddl2}, which is similar to STRIPS~\cite{fikes1971strips}.
$M = (F, A)$ is also referred to as the {\it model} in this work,
where
$F$ is the set of predicates used to specify the state and $A$ the set of actions used to update the state. 
Actions are associated with
a set of preconditions, add and delete effects. 
$I ,G$ are the initial and goal states, respectively.

\begin{definition}[Model Reconciliation~\cite{chakraborti2017plan}] A model reconciliation setting is a tuple $(\pi^*_{I, G}, \langle M^R , M^H \rangle)$ ($M^R \neq M^H$) under a given $I, G$, 
where $\pi^*_{I, G}$ corresponds to $\pi_{M^R}$ in Fig. \ref{fig1}
and represents the robot's behavior (plan) to be explained.
\end{definition}

Assuming rational agents, the $\pi^*_{I, G}$ above must satisfy $cost(\pi^{\ast}_{I,G},M^R) = cost^{\ast}_{M^R}(I,G)$, 
where $cost(\pi,M)$ returns the cost of a plan $\pi$ under the model $M$, and $cost^{\ast}_{M}(I,G)$ returns the cost of the optimal plan for the given initial and goal states under $M$. 
In other words, the robot's plan to be explained must be optimal under $M^R$.
It is assumed that the human obtains his expectation of the robot using $M^H$. Hence, when the robot's behavior does not match with the human's expectation, explanations must be made.
The goal of model reconciliation is to make the robot's plan $\pi^*_{I, G}$ also interpretable under the human's model $M^H$ (i.e., generable by $M^H$) by reducing the differences between $M^H$ and $M^R$. 

To define model differences, 
a mapping function $\Gamma$ was defined in ~\cite{chakraborti2017plan} to convert a planning problem into a set of features that fully specify the given problem.   
For simplicity, we modify the function here to remove the consideration of differences in the initial and goal states. As such, $\Gamma$ maps any planning problem from its model space $\mathcal{M}$ to the power set of its feature space $\mathcal{F}$ (i.e., $\Gamma: \mathcal{M} \longmapsto$ $2^\mathcal{F}$) as follows:

\vskip-10pt
\begin{equation*}
  \footnotesize
  \tau(f)=\begin{cases}
    a-has-precondition-f, & \text{if $f\in pre(a), a\in A$}.\\
    a-has-add-effect-f, & \text{if $f\in eff^+(a), a\in A$}.\\
    a-has-del-effect-f, & \text{if $f\in eff^-(a), a\in A$}.\\
    a-has-cost-f, & \text{if $f=c_a, a\in A$}.
  \end{cases}
\end{equation*}
\begin{align*}
\footnotesize
    \Gamma(M = (F, A)) = \{\tau(f)|\bigcup_{a\in A} \{f | f \in \{c_a\}\cup pre(a) \cup \\
    eff^+(a) \cup eff^-(a) \}  \}
\end{align*}

\begin{definition}[Explanation Generation~\cite{chakraborti2017plan}]
An explanation in a model reconciliation setting $(\pi^*_{I, G}, \langle M^R, M^H\rangle)$,
is a set of unit feature changes $\Delta$ to $M^H$ such that 
$1)$ $\Delta = \Gamma(\widehat{M^H}) \setminus \Gamma(M^H) \subseteq \Gamma({M^R})$, and
$2)$ $cost(\pi^*_{I, G}, {\widehat{M^H}}) - cost_{\widehat{M^H}}^*(I, G) < cost(\pi^*_{I, G}, {M^H}) - cost_{M^H}^*(I, G)$,
where $\widehat{M^H}$ is the model after the changes.
\label{def:exp}
\end{definition}

An explanation hence reconciles $M^R$ and $M^H$ by reducing their differences and making the cost difference between the human's expected plan $cost_{\widehat{M^H}}^*(I, G)$ and the robot's plan $cost(\pi^*_{I, G}, {\widehat{M^H}})$ smaller after the model updates. 
When the cost difference becomes $0$,
the robot's plan becomes optimal (and hence aligned with that) in the human's model.

\begin{definition}[Complete Explanation~\cite{chakraborti2017plan}]
An explanation is complete if it satisfies $cost(\pi^{\ast}_{I,G}, \widehat{M^H}) = cost^{\ast}_{\widehat{M^H}}(I,G)$.
\end{definition}

A minimal complete explanation (MCE) \cite{chakraborti2017plan} is defined as a complete explanation that 
contains the minimum number of unit feature changes.

\section{Online Explanation Generation (OEG)}

Prior work on explanation generation (including \cite{chakraborti2017plan}) focuses on providing the 
rationale behind the robot's decision making. 
It is often assumed that the explanation is provided in its entirety before the execution. 
As such, the cognitive requirement of the human for understanding the explanation is largely ignored:
when complex explanations are involved (such as in cognitively demanding tasks), communicating all information at the beginning becomes impractical.  
In such cases, it is desirable to communicate explanations in an incremental fashion. 
In this paper, we introduce \textit{online explanation generation} to address the above issue. 
The key idea here is to break up an explanation into multiple parts while ensuring consistency for interpretation, and communicate them separately during plan execution. 
Each part of an explanation is referred to as a {\it sub-explanation}.
A key observation that allows us to break up an explanation is that
only a part of the robot's plan needs to be explained by each sub-explanation given at a specific time step.
Next, we discuss three variations of OEG methods. 

\subsection{OEG for Matching Plan Prefix (OEG-PP)} \label{sps}

In this variation, each sub-explanation is required to explain a prefix of the robot's plan, such that it is 
consistent with the prefix of the human's expectation of the robot's plan. 
The sub-explanations are made incrementally in the sense that
each sub-explanation, when combined with the previous ones, explains a longer prefix of the robot's plan. 
The implication here is that the human's expected plan 
after all the sub-explanations will necessarily be the same
as the robot's plan, which is the longest prefix of itself. 

\begin{definition}[OEG-PP] 
An online explanation for matching plan prefix is a set of sub-explanations in the form of $\langle e_k, t_k \rangle$,
where $e_k$ represents the set of unit feature changes to be made as the $k$th sub-explanation {\it before} executing step $t_k$ (the $t_k$th action) of the plan, such that the following holds:
\begin{equation} \label{OnlineEq}
    \begin{multlined}[b]
        \hskip-2pt\forall k > 0, \textit{Prefix}(\pi^{\ast}_{I,G}, t_{k}-1) = \textit{Prefix}(\pi^{H}_{E_{k-1}}, t_{k}-1) 
        \\ s.t. \quad \Gamma(M^{H}_{E_{k-1}}) = \Gamma(M^H) \cup  E_{k-1},
        \\ 
       E_{k - 1} = \bigcup_1^{k - 1} e_i  \text{, and }
        E_{k - 1} \subseteq \Gamma(M^R)
    \end{multlined}
\end{equation}
\end{definition}
where \textit{Prefix}($\pi, t)$ returns the prefix of a plan $\pi$ up to step $t$ (inclusive). 
$E_{k}$ represents the union of all sub-explanations up to the $k$th sub-explanation and
$\pi^{H}_{E_{k}}$ the optimal plan created under $M^{H}_{E_k}$, which denotes $M^H$ after incorporating all the changes from $e_1$ to $e_k$.
More intuitively, at any step $k - 1$, the corresponding sub-explanation $e_{k - 1}$ is only responsible to explain the actions from $t_{k - 1}$ and onward until $t_k - 1$ in the robot's plan.

To generate each $\langle e_k, t_k \rangle$, the search process must consider how the sequence of model changes as a result of each sub-explanation would result in the change of the human's expectation.
This allows us to convert the problem of online explanation generation to the problem of model space search as in~\cite{chakraborti2017plan}.
The challenge here is that the model changes are not independent, i.e., future sub-explanations may 
have violated the condition in Eq. \eqref{OnlineEq} for the earlier sub-explanations. 
In such cases, an online explanation may become undesirable since 
the human may question the robot's earlier actions at a later stage, even though they appeared reasonable. This situation would introduce cognitive dissonance that may affect the human's understanding of the robot's plan.

To address this issue, it must be ensured that the model changes in the sub-explanations $e_k$ and onward, 
would not change the plan prefix that is already established up to plan step $t_k - 1$.
This can be achieved by searching backward from $M^R$ to $M^H$.
More specifically, given the model reconciliation setting for an explanation generation problem $(\pi_{I,G}^*, \langle M^R, M^H \rangle)$, the following process can be performed recursively to determine each sub-explanation.
First, we compute the human's expected plan using $M^H$, which is denoted as $\pi_H$. 
Denote the index of the first action where $\pi_{I,G}^*$ and $\pi_H$ differ as $t_1$, which is the timing for the first sub-explanation.  
To determine $e_1$, our search starts from $M^R$.
It finds the largest set of model changes to $M^R$, denoted as $\overline{e_1}$, such that {\it Prefix($\pi_{I,G}^*, t_1)$} = {\it Prefix($\pi^H_{M}, t_1)$} under any $M$ that lies in between $M^R \setminus \overline{e_1}$ and $M^R$ (i.e., 
$\Gamma(M^R) \setminus \overline{e_1} \subseteq \Gamma(M) \subseteq M^R$).
In this way, we guarantee that no change in $\overline{e_1}$ can violate the condition in Eq. \eqref{OnlineEq} once all the feature changes in $e_1$ are explained,
which is exactly what we strive for!
$e_1$ is then computed as the complement of $\overline{e_1}$, or $e_1 = \Gamma(M^R) \setminus (\Gamma(M^H) \cup \overline{e_1})$.
Now that we have found $\langle e_1, t_1 \rangle$, we can set $M^H$ to be $\Gamma(M^R) \setminus \overline{e_1}$, or equivalently $\Gamma(M^H) \cup e_1$ (i.e., $M_{E_1}^H$), to determine the next sub-explanation in a recursive manner. 
The recursion stops when the human's expected plan under $\Gamma(M^H) \cup \bigcup^{k}_{i=1} e_i$ (i.e., $M_{E_k}^H$) matches with $\pi_{I,G}^*$ for the first time,
where $e_k$ becomes the last sub-explanation. 

\begin{figure*}
\centering
    \includegraphics[width=0.79\textwidth]{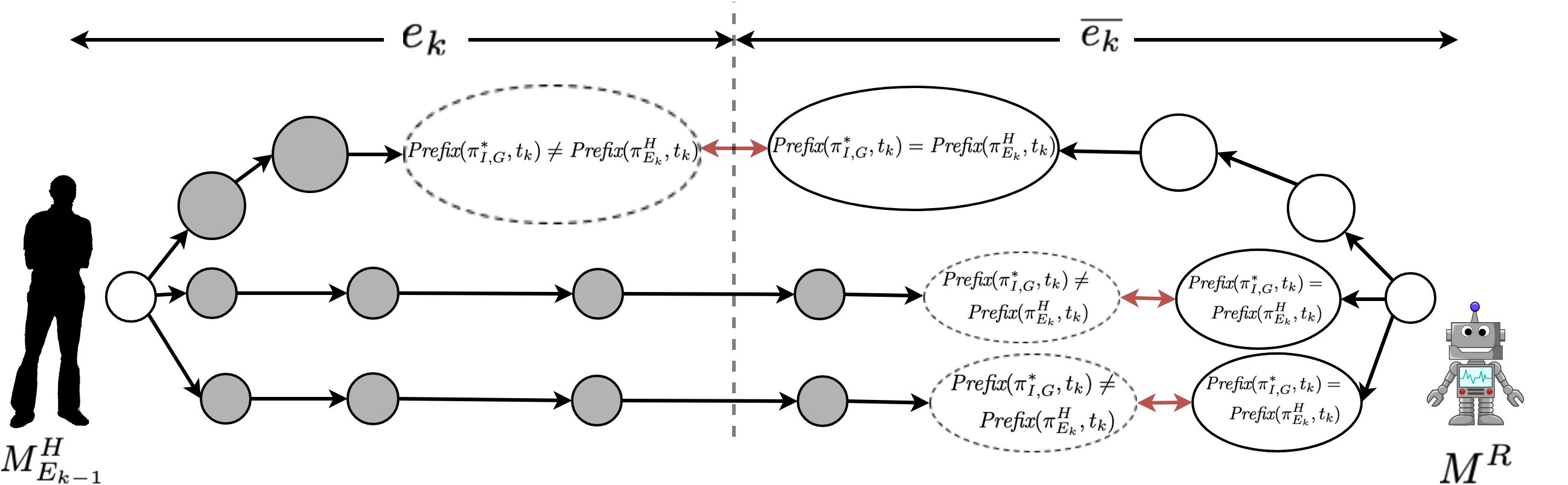}
    \caption{The model space search process for the $k$th sub-explanation in OEG-PP. 
    The search starts from $M^R$ (similar to that used for MCE in ~\cite{chakraborti2017plan}) until finding the largest set of $\overline{e_k}$ (or smallest $e_k$) that satisfies  {\it Prefix($\pi_{I,G}^*, t_k)$} = {\it Prefix($\pi^H_{E_k}, t_k)$}, under any $M$ that is in between $M^R \setminus \overline{e_k}$ and $M^R$. 
    Each node represents a candidate model and
    each edge a unit feature change. 
    The gray nodes are nodes that are not expanded in the search.}
    \label{search-oeg}
    \vspace{-5mm}
\end{figure*}

The model space search in OEG-PP for determining the $k$th sub-explanation is illustrated in Fig. \ref{search-oeg}. 
In practice, this search is computationally expensive. 
Hence, we implement an approximate method 
that searches forward from $M_{E_{k-1}}^H$ for 
the $k$th sub-explanation. 
The search is stopped when the smallest
$e_k$ that satisfies {\it Prefix($\pi_{I,G}^*, t_k)$} = {\it Prefix($\pi^H_{E_k}, t_k)$} is found. 
This approach is more efficient but comes with the cost that no guarantee can be made regarding the latter sub-explanations--they may introduce violations to the condition in Eq. \eqref{OnlineEq} for the earlier steps.
If this happens, we backtrack.
The implication here is that this method can no longer be used as an {\it online planning} method (i.e., computing the $e_k$'s online):
even though the sub-explanations are communicated online, they must be created offline. 

\subsection{OEG for Matching Next Action (OEG-NA)} \label{na}
In this variation, we relax the requirement in Eq. \eqref{OnlineEq} by requiring only the very next action to be interpretable at any step.
The assumption here is that the human would not  evaluate the robot's behavior retrospectively (or that its influence is minimal),
which is reasonable in cognitively demanding tasks where humans must focus more on the current situation due to a very limited cognitive span in such cases~\cite{paas2003cognitive}.
It is also worth noting that OEG-PP and OEG-NA
represent the two ends of the spectrum for online explanation generation where OEG-PP considers all actions occurred previously while OEG-NA ignores them all. It is expected that some method in between may work the best. Such analysis will be performed in our future work. 

\begin{definition}[OEG-NA] 
An online explanation for matching next action is a set of sub-explanations in the form of $\langle e_k, t_k \rangle$ such that the following is satisfied:
\begin{equation} \label{Relaxed}
    \begin{multlined}[b]
        \forall k > 0, 
        \pi^{\ast}_{I,G}[t_{k - 1}:t_k - 1] = \pi_{E_{k - 1}}^H[t_{k - 1}:t_k - 1] \\ 
        s.t. \quad \Gamma(M^{H}_{E_{k-1}}) = \Gamma(M^H) \cup  E_{k-1},
        \\ 
       E_{k-1} = \bigcup_1^{k-1} e_i \text{, and }
        E_{k-1} \subseteq \Gamma(M^R)
    \end{multlined}
\end{equation}
\end{definition}

The search for OEG-NA naturally starts from $M_{E_{k-1}}^H$ for $\langle e_k, t_k \rangle$
since we no longer worry about matching the prefix.

\subsection{OEG for Matching Any Prefix (OEG-AP)} \label{ap}
One assumption made in both OEG-PP and OEG-NA is that the optimal plan for a given $I, G$ pair is always unique. 
When we estimate the human's expected plan under a candidate model $M_{E_{k}}^H$ while searching for the $k$th sub-explanation, 
this assumption allows us to use the plan $\pi_{E_k}^H$ returned by any optimal planner, since they will always be the same.  
$\pi_{E_k}^H$ is then compared against $\pi_{I, G}^*$ to determine whether the $e_k$ (incorporated into $M_{E_k}^H$) satisfies the requirements of OEG.
When multiple optimal plans are present, 
the above check only needs to work for 
one of those plans.
In this variation, we relax the uniqueness assumption of the optimal plans.

\begin{definition}[OEG-AP] 
An online explanation for matching any prefix is a set of sub-explanations in the form of $\langle e_k, t_k \rangle$ such that the following is satisfied:
\begin{equation} \label{eq:indp}
    \begin{multlined}[b]
        \exists \pi_{E_{k-1}}^H \in \Pi_{E_{k-1}}^H \\
        \forall k > 0, \textit{Prefix}(\pi^{\ast}_{I,G}, t_{k}-1) = \textit{Prefix}(\pi^{H}_{E_{k-1}}, t_{k}-1) 
        \\ s.t. \quad \Gamma(M^{H}_{E_{k-1}}) = \Gamma(M^H) \cup  E_{k-1},
        \\ 
       E_{k - 1} = \bigcup_1^{k - 1} e_i \text{, and }
        E_{k - 1} \subseteq \Gamma(M^R)
    \end{multlined}
\end{equation}
\end{definition}
where $\Pi_{E_{k-1}}^H$ represents the set of all optimal plans under $M^H_{E_{k-1}}$.
A similar definition can be provided for OEG-NA after removing the uniqueness assumption.

To check for a candidate $e_k$, 
according to our previous discussion, 
we need to search for the largest set of model changes to $M^R$, denoted as $\overline{e_k}$,
such that {\it Prefix($\pi_{I,G}^*, t_k)$} = {\it Prefix($\pi^H_{E_k}, t_k)$}.
An obvious solution to OEG-AP is to obtain $\Pi_{E_{k}}^H$ by computing all the optimal plans under $M^H_{E_{k}}$. 
This approach however is computationally expensive. 
Instead, we implement a compilation approach. 
In this approach, to check the above condition,
we only need to solve two planning problems. 
The first planning problem is simple: finding an optimal plan $M^H_{E_{k}}$ under the given $I, G$. 
We denote the returned plan by any optimal planner as $\pi^H_{E_{k}}$ as usual.
The second one is trickier in which we need to obtain a problem under $M^H_{E_{k}}$ such that {\it any} optimal plan
would have to satisfy the condition {\it Prefix($\pi_{I,G}^*, t_k)$} = {\it Prefix($\pi^H_{E_k}, t_k)$}. 
We denote the plan returned as $\hat{\pi}^H_{E_{k}}$.
Now, we know that if the cost of $\hat{\pi}^H_{E_{k}}$ is equal to that of $\pi^H_{E_{k}}$, 
 there must exist an optimal plan in the human's model that matches the prefix of the robot's plan. 
Otherwise, no such plan exists and a sub-explanation must be made.
Hence, the key here is to ensure that a given plan prefix is always satisfied in a compiled model.

It turns out that this is not difficult to achieve. For all $ a_i, a_{i+1} \in $ {\it{Prefix}}$(\pi_{I, G}^*, t_k)$, where $a_i, a_{i+1}$ are two consecutive actions in $\pi_{I,G}^*$,  the compilation can be achieved by adding a predicate $p_i$ to $a_i$ as an effect, which is also added as a precondition for $a_{i+1}$. $a_{i+1}$, in its turn, adds $p_{i+1}$ as an effect which is a precondition for $a_{i+2}$, etc. 
The search process is the same as that described in OEG-PP. 
The search stops when any optimal plan in the human's updated model matches the robot's plan. 
In contrast to OEG-PP, the plan that is returned by an optimal planner under the human's model after an OEG-AP may not be exactly the robot's plan. 

\section{Evaluation}
\vskip-1pt
\begin{table*}
\centering
\begingroup
\setlength{\tabcolsep}{3.5pt} 
\renewcommand{\arraystretch}{1.2} 
    \centering
    \begin{tabular}{c|c|r|c|c|r|c|c|r|c|r}
    \hline
     Pr. & \multicolumn{2}{c|}{OEG-PP}  & \multicolumn{3}{c|}{OEG-NA} & \multicolumn{3}{c|}{OEG-AP} & \multicolumn{2}{c}{MCE}  \\
     ~       &      $\sum e_k/|e_k|$    &  Time &      $\sum e_k/|e_k|$ &  Dist.   & Time &   $\sum e_k/|e_k|$         &  Dist.    & Time &  $|E|$ & Time   \\ \hline \hline
      \multicolumn{11}{c}{Rover} \\ \cline{1-11} 
    P1             & 3/1.5      & 8.9    & 7/1.2  & 0.40 & 17.9    & 2/1.0             & 0.40        & 6.9      & 3  & 28.9  \\ 
    P2             & 5/1.7     & 22.3   & 7/1.4 & 0.11  & 42.6    & 3/1.0             & 0.11      & 18.3     & 5  & 150.5 \\ 
    P3             & 6/1.5      & 18.7   & 8/1.1  & 0.07 & 21.3    & 3/1.0             & 0.07      & 1.6      & 5  & 176.2 \\ 
    P4             & 6/1.5      & 51.0   & 8/1.3  & 0.13 & 94.8    & 5/1.3          & 0.13      & 45.4     & 6  & 314.2 \\ 
    P5             & 5/1.7     & 54.8   & 8/1.3 & 0.14 & 106.7   & 3/1.5             & 0.14      & 50.4     & 4  &  272.8 \\ 
\cline{1-11}
    \multicolumn{11}{c}{Barman} \\ \cline{1-11} 
    P1             & 5/1.3     & 43.0   & 5/1.3    & 0.91 & 59.9    & 2/1.0           & 0.94     & 24.4  & 5 & 180.0 \\ 
    P2             & 5/1.0        & 36.2   & 5/1.0    & 1.00 & 33.0    & 3/1.0           & 0.90     & 9.4  & 5  & 38.9 \\ 
    P3             & 5/1.3     & 36.8   & 5/1.0    & 0.90 & 46.8    & 3/1.5         & 0.71     & 9.7  & 5  & 51.8 \\ 
    P4             & 5/1.3     & 78.4   & 5/1.0  & 0.84 & 69.0    & 4/1.0           & 0.56     & 20.4  & 5 & 61.9 \\ 
    P5             & 5/1.7     & 41.9   & 5/1.0    & 0.89 & 54.7    & 3/1.5           & 0.56     & 10.2  & 5 & 61.5 \\ 
    \end{tabular}
\endgroup
    \caption{Comparison of explanation size, average sub-explanation size (for online only), plan distance between $\pi_{E_k}^H$ and $\pi_{I,G}^*$ (when applicable) and time (in seconds) using the different methods for the IPC Rover and Barman domains.
    }
    \label{simRes}
    \vspace{-2 em}
\end{table*}

We evaluate our methods for online explanation generation both synthetically and in simulation with human subjects, and compare them with variations of minimally complete explanations (MCE) \cite{chakraborti2017plan} as baselines. For the synthetic evaluation, 
our aim is to show how online explanations differ from MCEs. 
We evaluate our methods and MCE on $10$ different problems across the IPC Rover and Barman domains~\cite{IPC}. 
For human subject study, our aim is to verify the following: 
\begin{itemize}
    \item {\it Online explanations reduce mental workload and improve task performance.}
\end{itemize}
A modified rover domain (Sec. \ref{secsec:human-study}) is used.
In all evaluations, 
$M^R$ is the true domain model, 
and $M^H$ is created by removing model features from $M^R$. 
All results are collected on a $2.2$ GHz quad core Macbook Pro with $16$ GB RAM.

\subsection{Simulation Results} \label{simulation}
Table \ref{simRes} presents the simulation results comparing OEG-PP, OEG-NA and OEG-AP with MCE. 
The benefits of online explanations are clear:
the average size of sub-explanations is significantly smaller than the size of MCE, although the sum of their sizes is generally larger than the size of MCE.
This shows that most explanations can indeed be broken up and communicated incrementally while
subject to the requirements of online explanations!
The effect of OEG-AP on the size of explanations is interesting,
which suggests that removing the uniqueness assumption of the optimal plan has a positive impact on explanation generation: 
the sum of sub-explanations has a size that is smaller than MCE. 
This intuitively makes sense since not all the sub-explanations in MCE may be required as long as the robot's plan is optimal in the updated human's model (but differs from the plan found there by an optimal planner).  
To see the influence of removing the uniqueness assumption from another angle, 
for both OEG-NA and OEG-AP, we evaluate how the human's expected plan ($\pi_{E_k}^H$) after the explanation (returned by an optimal planner) may be different from $\pi_{I, G}^*$ using action plan distance,
which has a value between $0$ (no difference) and $1$ (maximum difference).
For OEG-NA, this distance is generally non-zero since only the very next action is considered when making a sub-explanation.
For OEG-AP, the distance is also non-zero in general but due to the non-uniqueness of the optimal plan. 
Computationally, OEG methods are generally a bit faster than MCE which may appear to be surprising. 
Some analysis reveals that this is due to the fact that the incremental search in online explanation generation in fact reduces the search space by removing candidate features to be added to $M_H$ for later searches. For OEG-NA and OEG-AP, this may also be due in part to the fact that they often terminate earlier and before 
$\pi_{E_k}^H$  becomes exactly $\pi_{I, G}^*$.

\subsection{Human Study}
\label{secsec:human-study}
\vskip-5pt
To test our hypothesis, 
we compare the explanations created by our methods with variations of MCE methods in a modified rover domain. 
The task is for the rover to collect and analyze soil and rock samples, take pictures of targets, and send them to the lander.
To ensure that the performance difference is not solely due to breaking up the information, we implement another baseline 
that randomly breaks up an MCE into multiple parts and communicates each part separately so that they are uniformly distributed through the plan execution (referred to as MCE-R). 

\begin{figure}
  \centering
    \includegraphics[width=0.87\columnwidth]{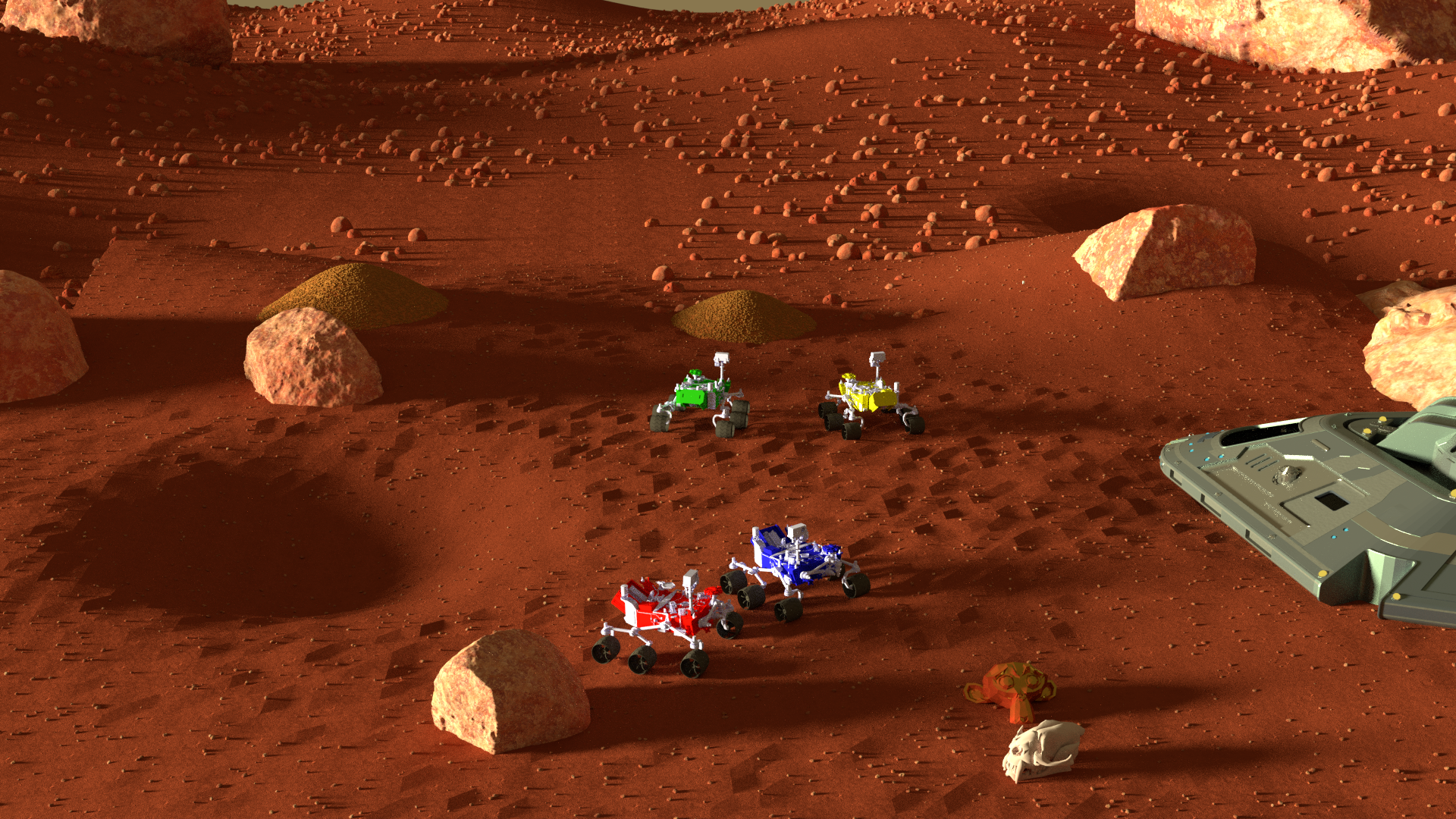}
  \caption{The 3D visualization of the modified rover domain. There are four robots on Mars, each has a different camera resolution and sampling equipment. The mission is to sample soil, rock and take images at different locations and communicate it to the lander shown on the right side of the picture.}
   \label{problem}
  \vskip-15pt
\end{figure}

We conducted our experiment using Amazon Mechanical Turk (MTurk) with a 3D simulation of the rover domain (see Fig. \ref{problem}).
The subjects were first given an introduction to the rover domain and the task they were supposed to help with. 
In the experiment, we deliberately removed certain information from the introduction. In particular,
we did not inform them that the storage space and  memory of the rover is limited, 
the camera must be calibrated, and calibrated with respect to the target before taking an image.
These introduced the 
differences between $M^H$ and $M^R$.

Each subject was given a 30-minute limit to finish the task. Explanations were provided using plain English language and the rover actions were depicted using GIF images in the 3D simulation as the rover executed the plan.
The human subject acted as the rover's supervisor,
and was asked to determine whether each of the rover's action was questionable or not.
Random actions were added into the plan to make sure that the subject must question some actions to perform well. 
Each subject was only allowed to perform the task for one setting (OEG-PP, OEG-NA, OEG-AP, MCE, or MCE-R) to reduce the influence of learning from repeated runs.
To simulate highly demanding tasks, we have  incorporated three spatial puzzles as secondary tasks. 
At the end of the study, the subjects were provided the NASA TLX to evaluate the workload~\cite{nasa} under several categories~\cite{tsang1996diagnosticity}. 

{\textbf{\textit{Results}}}:
We created the surveys using Qualtrics and recruited $150$ human subjects on MTurk, with $30$ subjects for each setting. 
To improve the quality of the responses, we set the criteria that the worker's HIT acceptance rate must be greater than $98\%$. 
After filtering out invalid responses (that failed to identify the $2$ purposely inserted random actions out of a total of $30$ actions in the plan), we obtained $94$ valid responses in total: $19$ for each of MCE-R and MCE, $20$ for OEG-PP, and $18$ for each of OEG-NA and OEG-AP. 
Their ages ranged from $18$ to $70$, and $29.8\%$ of them were female.

\begin{table}
\setlength{\tabcolsep}{4pt} 
    \footnotesize
    \centering
    \begin{tabular}{l|c|c|c|c|c|c}
    \hline
    ~               & MCE-R & MCE   & OEG-PP & OEG-NA & OEG-AP & Random \\ \hline
    Accuracy        & 0.746 & 0.804 & \textbf{0.858}  & \textbf{0.852}  & \textbf{0.872} &  \\ \hline
    \# Actions & 8.789 & 7.263 & \textbf{5.250}   & \textbf{5.330}   & \textbf{4.940} & 2.0/30  \\ \hline
    \end{tabular}
    \caption{The accuracy and number of questionable actions based on the subjects' feedback for the five settings. The ground truth for questionable actions is $2$ out of $30$ in total.}
    \label{tab:accuracy}
    \vskip-20pt
\end{table}

\begin{figure}
   \includegraphics[width=\columnwidth]{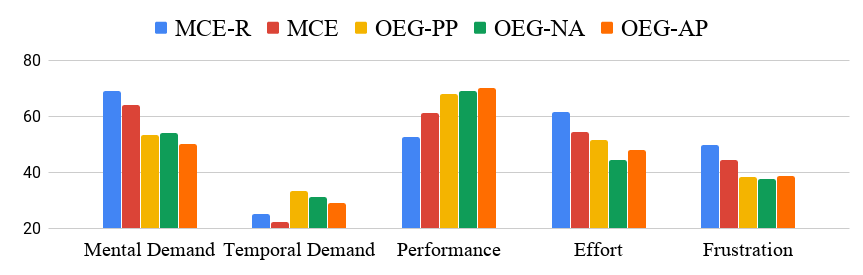}
   \caption{Comparison of TLX categories for the five settings.}
   \label{fig:chartTLX}
 \end{figure}

The results show that OEGs in general performed significantly better than the baselines in both objective (Table \ref{tab:accuracy}) and subjective measures (Fig. \ref{fig:chartTLX}). 
Table \ref{tab:accuracy} shows that the numbers of questionable actions are significantly lower for OEGs than MCEs (with $p$-values $< 0.001$). This indicates that the subjects had more trust towards robots in the OEG settings. The accuracy for identifying the correct actions (questionable vs. non-questionable) is also higher for OEGs (with $p$-values $< 0.001$). Among the three OEG methods, OEG-AP performed the best but no significant differences were observed in either the objective or subjective measures.
This seems to suggest that the performances were dominated mainly by the average size of sub-explanations, which did not vary much among the OEGs (i.e., $\sum_{e_k}/|e_k|$: OEG-PP 6/1.5, OEG-NA 5/1.25, OEG-AP 3/1.0, MCE 5/NA, MCE-R 5/1.0).

\begin{figure}
    \centering
    \vskip-12pt
    \includegraphics[width=0.7\columnwidth]{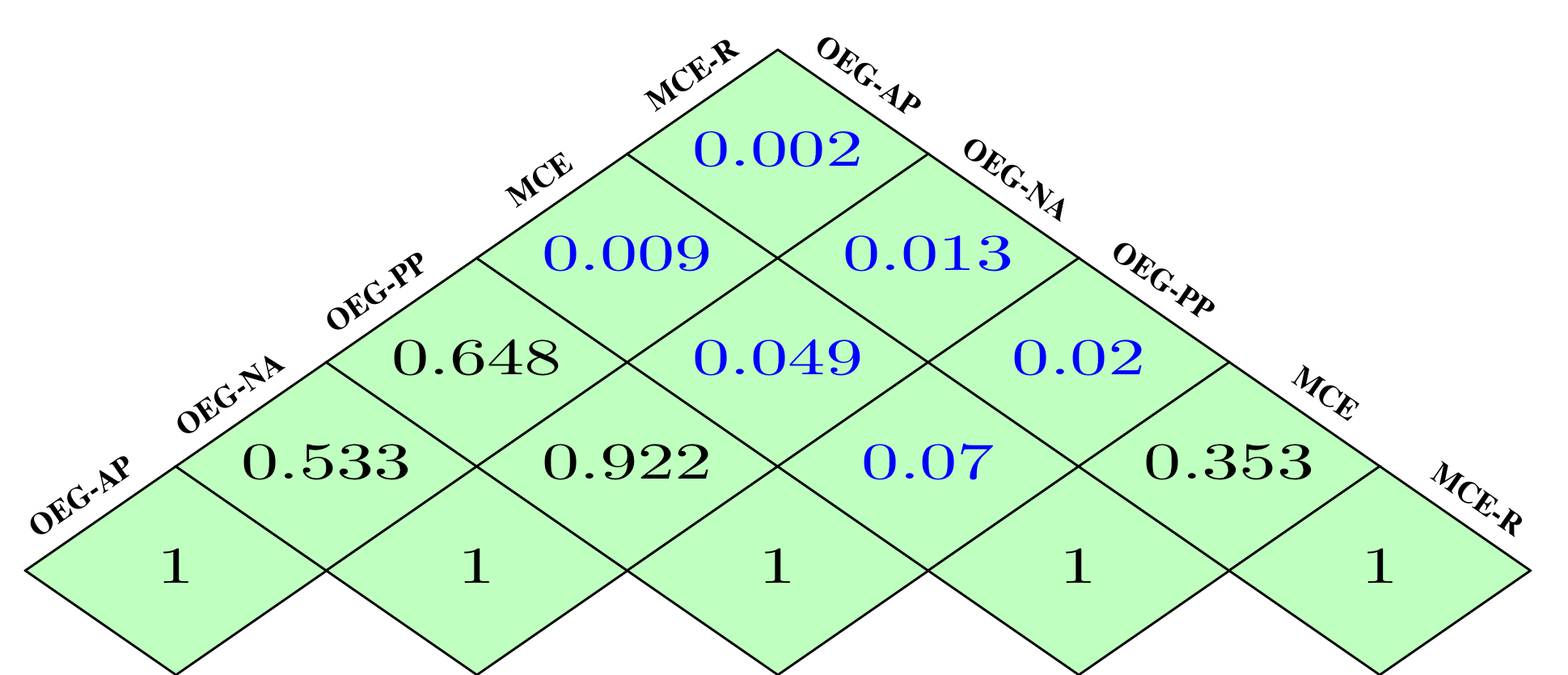}
    \vskip-3pt
    \caption{$p$-values for the weighted sum of the subjective measures, with weights $1.0$ for all TLX categories.  }
    \label{fig:pvalue}
    \vskip-20pt
\end{figure}

It is worth noting that MCE-R performed {\it worse} than MCE objectively with $p$-values $0.043$ and $0.028$ respectively for the two measures in Table \ref{tab:accuracy}, 
which suggests that the performance difference was unlikely due to simply breaking up the information, thus confirming the usefulness of OEGs. 
The subjective measures in Fig. \ref{fig:chartTLX} for the most part reaffirm the conclusions.
Due to intertwining explanations with plan execution, OEGs are expected to create more temporal demand.
The $p$-values for the subjective measures are presented in Fig. \ref{fig:pvalue}. 
The results indicate statistically significant differences between OEGs and MCEs. The group-wise $p$-value is $0.0068$ between OEGs and MCEs.

\section{Conclusions}
In this paper, we introduced a novel formulation for explanation generation that was focused on reducing the mental workload for the human to interpret an explanations. 
We took a step further from prior work, which considered only the correct explanations, by proposing explanations that were also easily understandable.
We provided three methods and evaluated them both in simulation and with human subjects. 
Results confirmed that they improved task performance and reduced mental workload.

\section*{ACKNOWLEDGMENT}
We thank the anonymous reviewers for their helpful comments. This research is supported in part by the NSF grant IIS-1844524, the NASA grant NNX17AD06G, and the AFOSR grant FA9550-18-1-0067

                                  
\bibliographystyle{IEEEtrans}
\bibliography{references}

\end{document}